\documentclass[11pt,a4paper]{article}
\usepackage[hyperref]{acl2020}
\usepackage{times}
\usepackage{latexsym}
\usepackage{xurl}
\usepackage{array}
\usepackage{graphicx}  
\usepackage{enumitem}
\usepackage{supertabular,booktabs}
\usepackage{xcolor}
\usepackage{color, colortbl}
\usepackage[explicit]{titlesec}
\usepackage{makecell}
\usepackage{multirow}
\usepackage{pifont}
\usepackage[noorphans,vskip=0ex]{quoting}

\definecolor{LGray}{gray}{0.9}
\definecolor{Gray}{gray}{0.8}
\definecolor{DGray}{gray}{0.7}

\aclfinalcopy

\title{Analyzing Political Parody in Social Media}

\author{
    Antonis Maronikolakis$^1$\Thanks{ Equal contribution.} \Thanks{ Work was done while at the University of Sheffield.} \quad {\bf Danae S\'{a}nchez Villegas$^{2*}$}  \\ {\bf Daniel Preo\c{t}iuc-Pietro$^3$} \quad {\bf Nikolaos Aletras$^2$}\\
    $^1$ Center for Information and Language Processing, LMU Munich, Germany\\
    $^2$ Computer Science Department, University of Sheffield, UK\\
    $^3$ Bloomberg\\
    {\small
    {\tt antmarakis@cis.lmu.de}, {\tt \{dsanchezvillegas1, n.aletras\}@sheffield.ac.uk}}\\
    {\small
    {\tt dpreotiucpie@bloomberg.net}}
}
    
\date{}

\begin{document}
\maketitle
\begin{abstract}




Parody is a figurative device used to imitate an entity for comedic or critical purposes and represents a widespread phenomenon in social media through many popular parody accounts. In this paper, we present the first computational study of parody. We introduce a new publicly available data set of tweets from real politicians and their corresponding parody accounts. We run a battery of supervised machine learning models for automatically detecting parody tweets with an emphasis on robustness by testing on tweets from accounts unseen in training, across different genders and across countries. Our results show that political parody tweets can be predicted with an accuracy up to 90\%. Finally, we identify the markers of parody through a linguistic analysis. Beyond research in linguistics and political communication, accurately and automatically detecting parody is important to improving fact checking for journalists and analytics such as sentiment analysis through filtering out parodical utterances.\footnote{Data is available here: \url{https://archive.org/details/parody_data_acl20}}
 

\end{abstract}

\section{Introduction}
\label{sec:intro}




Parody is a figurative device which is used to imitate and ridicule a particular target~\cite{Rose1993} and has been studied in linguistics as a figurative trope distinct to irony and satire~\citep{Kreuz1993,Rossenknill1997}. Traditional forms of parody include editorial cartoons, sketches or articles pretending to have been authored by the parodied person.\footnote{The `Kapou Opa' column by K. Maniatis parodying Greek popular persons was a source of inspiration for this work - \url{https://www.oneman.gr/originals/to-imerologio-karantinas-tou-dimitri-koutsoumpa/}} A new form of parody recently emerged in social media, and Twitter in particular, through accounts that impersonate public figures. \citet{Highfield2016} defines parody accounts acting as: \emph{a known, real person, for obviously comedic purposes. There should be no risk of mistaking their tweets for their subject's actual views; these accounts play with stereotypes of these figures or juxtapose their public image with a very different, behind-closed-doors persona}.

A very popular type of parody is political parody which plays an important role in public speech by offering irreverent interpretations of political personas~\cite{Hariman2008}. Table~\ref{t:tweets_example} shows examples of very popular (over 50k followers) and active (thousands of tweets sent) political parody accounts on Twitter. Sample tweets show how the style and topic of parody tweets are similar to those from the real accounts, which may pose issues to automatic classification.

While closely related figurative devices such as irony and sarcasm have been extensively studied in computational linguistics~\cite{wallace2015computational,joshi2017automatic}, parody yet to be explored using computational methods. In this paper, we aim to bridge this gap and conduct, for the first time, a systematic study of political parody as a figurative device in social media. To this end, we make the following contributions:
\begin{enumerate}[noitemsep,topsep=0pt,leftmargin=1.3em]
    \item A novel classification task where we seek to automatically classify real and parody tweets. For this task, we create a new large-scale publicly available data set containing a total of 131,666 English tweets from 184 parody accounts and corresponding real accounts of politicians from the US, UK and other countries (Section~\ref{sec:data});
    \item Experiments with feature- and neural-based machine learning models for parody detection, which achieve high predictive accuracy of up to 89.7\% F1. These are focused on the robustness of classification, with test data from: a) users; b) genders; c) locations; unseen in training (Section~\ref{sec:results});
    \item Linguistic analysis of the markers of parody tweets and of the model errors (Section~\ref{sec:analysis}).
\end{enumerate}

\renewcommand*{\arraystretch}{1.2}
\begin{table*}[!t]
    \footnotesize
    \centering
    \begin{tabular}{|l|l|m{10cm}|}
        \hline
        \rowcolor[HTML]{9B9B9B}
        Account type & Twitter Handle & Sample tweet  \\
        \hline
        Real & \texttt{@realDonaldTrump} & The Republican Party, and me, had a GREAT day yesterday with respect to the phony Impeachment Hoax, \& yet, when I got home to the White House \& checked out the news coverage on much of television, you would have no idea they were reporting on the same event. FAKE \& CORRUPT NEWS! \\
        \rowcolor{LGray}
        Parody & \texttt{@realDonaldTrFan} & Lies! Kampala Harris says my crimes are committed in plane site! She’s lying! My crimes are ALWAYS hidden! ALWAYS!! \\

\hline \hline

        Real & \texttt{@BorisJohnson} & Our NHS will never be on the table for any trade negotiations. We’re investing more than ever before - and when we leave the EU, we will introduce an Australian style, points-based immigration system so the NHS can plan for the future. \\
        \rowcolor{LGray}
        Parody & \texttt{@BorisJohnson\_MP} & People seem to be ignoring the many advantages of selling off the NHS, like the fact that hospitals will be far more spacious once poor people can't afford to use them.  \\
        \hline
    \end{tabular}
    \caption{Two examples of Twitter accounts of politicians and their corresponding parody account with a sample tweet from each.}
    \label{t:tweets_example}
\end{table*}


We argue that understanding the expression and use of parody in natural language and automatically identifying it are important to applications in computational social science and beyond. Parody tweets can often be misinterpreted as facts even though Twitter only allows parody accounts if they are explicitly marked as parody\footnote{Both the profile description and account name need to mention this -- \url{https://help.twitter.com/en/rules-and-policies/parody-account-policy}} and the poster does not have the intention to mislead. For example, the Speaker of the US House of Representatives, Nancy Pelosi, falsely cited a Michael Flynn parody tweet;\footnote{\url{https://tinyurl.com/ybbrh74g}} and many users were fooled by a Donald Trump parody tweet about `Dow Joans'.\footnote{\url{https://tinyurl.com/s34dwgm}} Thus, accurate parody classification methods can be useful in downstream NLP applications such as automatic fact checking~\cite{Vlachos2014} and rumour verification~\cite{Karmakharm2019}, sentiment analysis~\cite{pang2008opinion} or nowcasting voting intention~\cite{Tumasjan2010,lampos-etal-2013-user,Tsakalidis2018}. 

Beyond NLP, parody detection can be used in: (i) political communication, to study and understand the effects of political parody in the public speech on a large scale~\cite{Hariman2008,Highfield2016}; (ii) linguistics, to identify characteristics of figurative language~\cite{Rose1993,Kreuz1993,Rossenknill1997}; (iii) network science, to identify the adoption and diffusion mechanisms of parody~\cite{vosoughi2018spread}.

\section{Related Work}


\paragraph{Parody in Linguistics}~Parody is an artistic form and literary genre that dates back to Aristophanes in ancient Greece who parodied argumentation styles in \textit{Frogs}. Verbal parody was studied in linguistics as a figurative trope distinct to irony and satire~\citep{Kreuz1993,Rossenknill1997} and researchers long debated its definition and theoretic distinctions to other types of humor~\cite{grice1975logic,sperber1984verbal,wilson2006pragmatics,dynel2014isn}. In general, verbal parody involves a highly situated, intentional, and conventional speech act~\cite{Rossenknill1997} composed of both a negative evaluation and a form of pretense or echoic mention~\cite{sperber1984verbal,wilson2006pragmatics,dynel2014isn} through which an entity is mimicked or imitated with the goal of criticizing it to a comedic effect. Thus, imitative composition for amusing purpose is an an inherent characteristic of parody \cite{franke1971note}. The parodist intentionally re-presents the object of the parody and flaunts this re-presentation~\cite{Rossenknill1997}.


\paragraph{Parody on Social Media}~Parody is considered an integral part of Twitter~\cite{vis2013twitter} and previous studies on parody in social media focused on analysing how these accounts contribute to topical discussions~\citep{Highfield2016} and the relationship between identity, impersonation and authenticity~\citep{Page2014}. Public relation studies showed that parody accounts impact organisations during crises while they can become a threat to their reputation~\citep{Wan2015}.

\paragraph{Satire}~Most related to parody, satire has been tangentially studied as one of several prediction targets in NLP in the context of identifying disinformation~\cite{Mchardy2019,de2019deciding}. \citep{rashkin-etal-2017-truth} compare the language of real news with that of satire, hoaxes, and propaganda to identify linguistic features of unreliable text. They demonstrate how stylistic characteristics can help to decide the text's veracity. The study of parody is therefore relevant to this topic, as satire and parodies are classified by some as a type of disinformation with `no intention to cause harm but has potential to fool'~\cite{wardle2018thinking}.

\paragraph{Irony and Sarcasm}~There is a rich body of work in NLP on identifying irony and sarcasm as a classification task~\cite{wallace2015computational,joshi2017automatic}. \citet{van-hee-etal-2018-semeval} organized two open shared tasks. The first aims to automatically classify tweets as ironic or not, and the second is on identifying the type of irony expressed in tweets. However, the definition of irony is usually `a trope whose actual meaning differs from what is literally enunciated'~\cite{van-hee-etal-2018-semeval}, following the Gricean belief that the hallmark of irony is to communicate the opposite of the literal meaning~\cite{wilson2006pragmatics}, violating the first maxim of Quality~\cite{grice1975logic}. In this sense, irony is treated in NLP in a similar way as sarcasm~\cite{gonzalez-ibanez-etal-2011-identifying,khattri-etal-2015-sentiment,joshi2017automatic}. In addition to the words in the utterance, further using the user and pragmatic context is known to be informative for irony or sarcasm detection in NLP~\cite{bamman2015contextualized,wallace2015computational}. For instance, \citet{oprea-magdy-2019-exploring} make use of user embeddings for textual sarcasm detection. In the design of our data splits, we aim to limit the contribution of this aspects from the results.



\paragraph{Relation to other NLP Tasks} The pretense aspect of parody relates our task to a few other NLP tasks. In authorship attribution, the goal is to predict the author of a given text~\cite{stamatatos2009survey,juola2008authorship,koppel2009computational}. However, there is no intent for the authors to imitate the style of others and most differences between authors are in the topics they write about, which we aim to limit by focusing on political parody. Further, in our setups, no tweets from an author are in both training and testing to limit the impact of terms specific to a particular person.

Pastiche detection~\cite{dinu2012pastiche} aims to distinguish between an original text and a text written by someone aiming to imitate the style of the original author with the goal of impersonating. Most similar in experimental setup to our task, ~\citet{preotiuc-pietro-devlin-marier-2019-analyzing} aim to distinguish between tweets published from the same account by different types of users: politicians or their staff. While both pastiches and staff writers aim to present similar content with similar style to the original authors, the texts lack the humorous component specific of parodies.

A large body of related NLP work has explored the inference of user characteristics. Past research studied predicting the type of a Twitter account, most frequently between individual or organizational, using linguistic features~\cite{de2012unfolding,mccorriston2015organizations,mac2017demographic}. A broad literature has been devoted to predicting personal traits from language use on Twitter, such as gender~\cite{burger2011discriminating}, age~\cite{nguyen2011author}, geolocation~\cite{cheng2010you}, political preference~\cite{volkova2014inferring,preoctiuc2017beyond}, income~\cite{preoctiuc2015studying,aletras2018predicting}, impact~\cite{lampos2014predicting}, socio-economic status~\cite{Lampos2016}, race~\cite{preoctiuc2018user} or personality~\cite{schwartz2013personality,preotiuc2016studying}.

\section{Task \& Data}
\label{sec:data}

We define parody detection in social media as a binary classification task performed at the social media post level. Given a post $T$, defined as a sequence of tokens $T=\{t_1,...,t_n\}$, the aim is to label $T$ either as parody or genuine. Note that one could use social network information but this is out of the paper's scope as we only focus on parody as a linguistic device.

We create a new publicly available data set to study this task, as no other data set is available. We perform our analysis on a set of users from the same domain (politics) to limit variations caused by topic. We first identify real and parody accounts of politicians on Twitter posting in English from the United States of America (US), the United Kingdom (UK) and other accounts posting in English from the rest of the world. We opted to use Twitter because it is arguably the most popular platform for politicians to interact with the public or with other politicians~\cite{Parmelee2011}. For example, 67\% of prospective parliamentary candidates for the 2019 UK general election have an active Twitter account.\footnote{\url{https://www.mpsontwitter.co.uk/}} Twitter also allows to maintain parody accounts, subject to adding explicit markers in both the user bio and handle such as \texttt{parody}, \texttt{fake}.\footnote{\url{https://help.twitter.com/en/rules-and-policies/parody-account-policy}} Finally, we label tweets as parody or real, depending on the type of account they were posted from. We highlight that we are not using user description or handle names in prediction, as this would make the task trivial.


\subsection{Collecting Real and Parody Politician Accounts}


We first query the public Twitter API using the following terms: \{\texttt{parody, \#parody, parody account, fake, \#fake, fake account, not real}\} to retrieve candidate parody accounts according to Twitter's policy. From that set, we exclude any accounts matching \texttt{fan} or \texttt{commentary} in their bio or account name since these are likely to be not posting parodical content. We also exclude private and deactivated accounts and accounts with a majority of non-English tweets.

After collecting this initial set of parody candidates, the authors of the paper manually inspected up to the first ten original tweets from each candidate to identify whether an account is a parody or not following the definition of a public figure parody account from~\citet{Highfield2016} (see Section~\ref{sec:intro}), further filtering out non-parody accounts. We keep a single parody account in case of multiple parody accounts about the same person. Finally, for each remaining account, the authors manually identified the corresponding real politician account to collect pairs of real and parody. 

Following the process above, we were able to identify parody accounts of 103 unique people, with 81 having a corresponding real account. 
The authors also identified the binary gender and location (country) of the accounts using publicly available records. This resulted in 21.6\% female accounts (women parliamentarians percentages as of 2017: 19\% US, 30\% UK, 28.8\% OECD average).\footnote{\url{https://data.oecd.org/inequality/women-in-politics.htm}} The majority of the politicians are located in the US (44.5\%) followed by the UK (26.7\%) while 28.8\% are from the rest of the world (e.g. Germany, Canada, India, Russia).




\subsection{Collecting Real and Parody Tweets}

We collect all of the available original tweets, excluding retweets and quoted tweets, from all the parody and real politician accounts.\footnote{Up to maximum 3200 tweets/account according to Twitter API restrictions.}
We further balance the number of tweets in a real -- parody account pair in order for our experiments and linguistic analysis not to be driven by a few prolific users or by imbalances in the tweet ratio for a specific pair. We keep a ratio of maximum $\pm 20\%$ between the real and parody tweets per pair by keeping all tweets from the less prolific account and randomly down-sampling from the more prolific one. Subsequently, for the parody accounts with no corresponding real account, we sample a number of tweets equal to the median number of tweets for the real accounts. Finally, we label tweets as parody or real, depending on the type of account they come from. In total, the data set contains 131,666 tweets, with 65,710 real and 65,956 parody.

\subsection{Data Splits}

To test that automatically predicting political parody is robust and generalizes to held-out situations not included in the training data, we create the following three data splits for running experiments:

\paragraph{Person Split}~We first split the data by adding all tweets from each real -- parody account pair to a single split, either train, development or test. To obtain a fairly balanced data set without pairs of accounts with a large number of tweets dominating any splits, we compute the mean between real and parody tweets for each account, and stratify them, with pairs of proportionally distributed means across the train, development, and test sets (see Table~\ref{tab:person_split}).



\begin{table}[!t]
\centering
\small
\resizebox{0.48\textwidth}{!}{
\begin{tabular}{lrrrrc}
\hline
\rowcolor[HTML]{C0C0C0}
\multicolumn{6}{c}{\textbf{Person}} \\
\hline
  & \textbf{Train} & \textbf{Dev} & \textbf{Test} & \textbf{Total} & \begin{tabular}[c]{@{}c@{}c@{}}\textbf{Avg.} \textbf{tokens} \\\textbf{(Train)}\end{tabular} \\ \toprule
Real      & 51,460          & 6,164         & 8,086          & 65,710     & 23.33  \\ \hline
Parody    & 51,706          & 6,164         & 8,086          & 65,956     & 20.15 \\ \hline
All     & 103,166         & 12,328        & 16,172         & 131,666    &  22.55 \\
\bottomrule


\end{tabular}
}
\caption{Data set statistics with the person split.}
\label{tab:person_split}
\end{table}

\paragraph{Gender Split}~We also split the data by the gender of the politicians into training, development and test, obtaining two versions of the data: (i) one with female accounts in train/dev and male in test; and (ii) male accounts in train/dev and female in test (see Table~\ref{tab:gender_split}).

\begin{table}[!t]
\resizebox{0.48\textwidth}{!}{%
\begin{tabular}{clrrr}
\rowcolor[HTML]{C0C0C0}
\multicolumn{5}{c}{\textbf{Gender}} \\
\hline
\textbf{Trained on}              & \multicolumn{1}{c}{} & \textbf{Real} & \textbf{Parody} & \textbf{Total} \\ 
\toprule
\multirow{3}{*}{\textbf{Female}} & \textbf{Train}                     & 10,081         & 11,036           & 21,117          \\ \cline{2-5} 
                                 & \textbf{Dev}                       & 302           & 230             & 532            \\ \cline{2-5} 
                                 & \textbf{Test (Male)}               & 55,327         & 54,690           & 110,017         \\ \hline
\multirow{3}{*}{\textbf{Male}}   & \textbf{Train}                     & 51,048         & 50,184           & 101,232         \\ \cline{2-5} 
                                 & \textbf{Dev}                       & 4,279          & 4,506            & 8,785           \\ \cline{2-5} 
                                 & \textbf{Test (Female)}             & 10,383         & 11,266           & 21,649          \\ 
                                 \bottomrule
\end{tabular}
}
\caption{Data set statistics with the gender split (Male, Female).}
\label{tab:gender_split}
\end{table}

\paragraph{Location split}~Finally, we split the data based on the location of the politicians. We group the accounts in three groups of locations: US, UK and the rest of the world (\textbf{RoW}). We obtain three different splits, where each group makes up the test set and the other two groups make up the train and development set (see Table~\ref{tab:location_split}). 

\begin{table}[!t]
\resizebox{0.48\textwidth}{!}{%
\begin{tabular}{llrrr}
\hline
\rowcolor[HTML]{C0C0C0}
\multicolumn{5}{c}{\textbf{Location}} \\
\hline
\textbf{Trained on}                   & \textbf{}             & \textbf{Real} & \textbf{Parody} & \textbf{Total} \\ \hline
\toprule
\multirow{3}{*}{\textbf{US \& RoW}} & \textbf{Train}        & 47,018        & 45,005          & 92,023         \\ \cline{2-5} 
                                      & \textbf{Dev}          & 1,030         & 2,190           & 3,220          \\ \cline{2-5} 
                                      & \textbf{Test (UK)}    & 17,662        & 18,761          & 36,423         \\ \hline
\multirow{3}{*}{\textbf{UK \& RoW}} & \textbf{Train}        & 33,687        & 35,371          & 69,058         \\ \cline{2-5} 
                                      & \textbf{Dev}          & 1,030         & 1,274           & 2,304          \\ \cline{2-5} 
                                      & \textbf{Test (US)}    & 30,993        & 29,311          & 60,304         \\ \hline
\multirow{3}{*}{\textbf{US \& UK}}    & \textbf{Train}        & 43,211        & 42,597          & 85,808         \\ \cline{2-5} 
                                      & \textbf{Dev}          & 5,444         & 5,475           & 10,919         \\ \cline{2-5} 
                                      & \textbf{Test (RoW)} & 17,055        & 17,884          & 34,939         \\ 
\bottomrule
\end{tabular}
}
\caption{Data set statistics with the location split (US, UK, Rest of the World--RoW).}
\label{tab:location_split}
\end{table}





\subsection{Text Preprocessing}
We preprocess text by lower-casing, replacing all URLs and anonymizing all mentions of usernames with placeholder token. We preserve emoticons and punctuation marks and replace tokens that appear in less than five tweets with a special `unknown' token. We tokenize text using DLATK \cite{schwartz-etal-2017-dlatk}, a Twitter-aware tokenizer. 

\section{Predictive Models}
\label{sec:models}

We experiment with a series of approaches to classification of parody tweets, ranging from linear models, neural network architectures and pre-trained contextual embedding models. Hyperparameter selection is included in the Appendix. 

\subsection{Linear Baselines}


\paragraph{LR-BOW}~As a first baseline, we use a logistic regression with standard bag-of-words (LR-BOW) representation of the tweets.  

\paragraph{LR-BOW+POS}~We extend LR-BOW using syntactic information from Part-Of-Speech (POS) tags. We first tag all tweets in our data using the NLTK tagger and then we extract bag-of-words features where each unigram consists of a token with its associated POS tag.

\subsection{BiLSTM-Att}
The first neural model is a bidirectional Long-Short Term Memory (LSTM) network \cite{hochreiter1997long} with a self-attention mechanism (BiLSTM-Att; \citet{zhou-etal-2016-attention}). Tokens $t_i$ in a given tweet $T=\{t_1,...,t_n\}$  are mapped to embeddings and passed through a bidirectional LSTM. A single tweet representation ($h$) is computed as the sum of the resulting contextualized vector representations ($\sum_ia_ih_i$) where $a_i$ is the self-attention score in timestep $i$. The tweet representation ($h$) is subsequently passed to the output layer using a sigmoid activation function.


\subsection{ULMFit} 
The Universal Language Model Fine-tuning (ULMFit) is a method for efficient transfer learning~\citep{ulmfit}. The key intuition is to train a text encoder on a language modelling task (i.e. predicting the next token in a sequence) where data is abundant, then fine-tune it on a target task where data is more limited. During fine-tuning, ULMFit uses gradual layer unfreezing to avoid catastrophic forgetting. We experiment with using AWD-LSTM~\cite{awd_lstm} as the base text encoder pretrained on the Wikitext 103 data set and we fine-tune it on our own parody classification task. For this purpose, after the AWS-LSTM layers, we add a fully-connected layer with a ReLU activation function followed by an output layer with a sigmoid activation function. Before each of these two additional layers, we perform batch normalization.



\subsection{BERT and RoBERTa} 
Bidirectional Encoder Representations from Transformers (BERT) is a language model based on transformer networks~\cite{Vaswani2017} pre-trained on large corpora~\cite{bert}. The model makes use of multiple multi-head attention layers to learn bidirectional embeddings for input tokens. It is trained for masked language modelling, where a fraction of the input tokens in a given sequence are masked and the task is to predict a masked word given its context. BERT uses wordpieces which are passed through an embedding layer and get summed together with positional and segment embeddings. The former introduce positional information to the attention layers, while the latter contain information about the location of a segment. Similar to ULMFit, we fine-tune the BERT-base model for predicting parody tweets by adding an output dense layer for binary classification and feeding it with the `classification' token. 


We further experiment with RoBERTa \cite{roberta}, which is an extenstion of BERT trained on more data and different hyperparameters. RoBERTa has been showed to improve performance in various benchmarks compared to the original BERT \cite{roberta}.

\subsection{XLNet} 
XLNet is another pre-trained neural language model based on transformer networks~\cite{xlnet}. XLNet is similar to BERT in its structure, but is trained on a permutated (instead of masked) language modelling task. During training, sentence words are permuted and the model predicts a word given the shuffled context. We also adapt XLNet for predicting parody, similar to BERT and ULMFit. 



\subsection{Model Hyperparameters}

We optimize all model parameters on the development set for each data split (see Section~\ref{sec:data}).

\paragraph{Linear models} For the LR-BOW, we use n-grams with $n$ = (1, 2), $n \in$ \{(1, 1), (1, 2), (1, 3) weighted by TF.IDF. For the LR-BOW+POS, we use TF with POS n-grams where $n =$ (1, 3). For both baselines we use L2 regularization.

\paragraph{BiLSTM-Att} We use 200-dimensional GloVe embeddings \cite{pennington2014glove} pre-trained on Twitter data. The maximum sequence length is set to 50 covering 95\% of the tweets in the training set. The LSTM size is $h$ = 300 where $h \in \{50,100,300\}$ with dropout $d$ = 0.5 where $d \in \{.2,.5\}$. We use Adam \cite{kingma2014adam} with default learning rate, minimizing the binary cross-entropy using a batch size of 64 over 10 epochs with early stopping.

\paragraph{ULMFit} We first update only the AWD-LSTM weights with a learning rate $l$ = 2e-3 for one epoch where $l \in $ \{1e-3, 2e{-3}, 4e{-3}\} for language modeling. Then, we update both the AWD-LSTM and embedding weights for one more epoch, using a learning rate of $l$ = 2e{-5} where $l \in $ \{1e{-4}, 2e{-5}, 5e{-5}\}. The size of the intermediate fully-connected layer (after AWD-LSTM and before the output) is set by default to $50$. Both in the intermediate and output layers we use default dropout of $0.08$ and $0.1$ respectively from \citet{ulmfit}.

\paragraph{BERT and RoBERTa} For BERT, we used the base model (12 layers and 110M total parameters) trained on lowercase English. We fine-tune it for 1 epoch with a learning rate $l$ = 5e{-5}  where $l \in$ \{2e{-5}, 3e{-5}, 5e{-5}\} as recommended in \citet{bert} with a batch size of $128$. For RoBERTa, we use the same fine-tuning parameters as BERT.

\paragraph{XLNet}~We use the same parameters as BERT except for the learning rate, which we set at $l$ = 4e{-}5 where $l \in$ \{2e{-5}, 4e{-5}, 5e{-5}\}.

\renewcommand{\arraystretch}{1.1}
\begin{center}
\begin{table*}[t!]
\centering
\begin{tabular}{|
>{\columncolor[HTML]{EFEFEF}}l |c|c|c|c|c|}
\hline
\multicolumn{6}{|c|}{\cellcolor[HTML]{C0C0C0}\textbf{Person}}\\
\hline
\multicolumn{1}{|c|}{\cellcolor[HTML]{C0C0C0}\textbf{Model}} & \cellcolor[HTML]{C0C0C0}\textbf{Acc} & \cellcolor[HTML]{C0C0C0}\textbf{P} & \cellcolor[HTML]{C0C0C0}\textbf{R} & \cellcolor[HTML]{C0C0C0}\textbf{F1} & \cellcolor[HTML]{C0C0C0}\textbf{AUC} \\ \toprule
LR-BOW                                                      & 73.95 $\pm$0.00                            & 70.08 $\pm$ 0.01                             & 83.53 $\pm$0.02                          & 76.19 $\pm$0.00                            & 73.96 $\pm$0.00                       \\ 
LR-BOW+POS                                                 & 74.33 $\pm$0.00                            & 71.34 $\pm$0.00                             & 81.19 $\pm$0.00                          & 75.95 $\pm$0.00                            & 74.34 $\pm$0.00                       \\ \hline
BiLSTM-Att                                                  & 79.92 $\pm$0.01                            & 81.63 $\pm$0.01                             & 77.11 $\pm$0.03                          & 79.29 $\pm$0.02                            & 79.91 $\pm$0.01                       \\ 
ULMFit                                                       & 81.11 $\pm$0.38                            & 75.57 $\pm$2.03                             & 84.97 $\pm$0.87                          & 81.05 $\pm$0.42                            & 81.10 $\pm$0.38                       \\
BERT                                                         & 87.65 $\pm$0.29                            & 87.63 $\pm$0.58                             & 87.67 $\pm$0.40                          & 87.65 $\pm$0.18                            & 87.65 $\pm$0.32                       \\ 
RoBERTa                                                      & \textbf{90.01 $\pm$0.35}                   & \textbf{90.90 $\pm$0.55}                    & \textbf{88.45 $\pm$0.22}                 & \textbf{89.66 $\pm$0.33}                   & \textbf{90.05 $\pm$0.29}              \\ 
XLNet                                                   & 86.45 $\pm$0.41                            & 88.24 $\pm$0.52                             & 85.18 $\pm$0.40                         & 86.68 $\pm$0.37                            & 86.45 $\pm$0.36                       \\ 
 \bottomrule
\end{tabular}
\caption{Accuracy (Acc), Precision (P), Recall (R), F1-Score (F1) and ROC-AUC for parody prediction splitting by person ($\pm$ std. dev.). Best results are in bold.}
\label{tab:results_person}
\end{table*}
\end{center}

\section{Results}
\label{sec:results}

This section contains the experimental results obtained on all three different data splits proposed in Section~\ref{sec:data}. We evaluate our methods (Section~\ref{sec:models}) using several metrics, including accuracy, precision, recall, macro F1 score, and Area under the ROC (AUC). We report results over three runs using different random seeds and we report the average and standard deviation.

\subsection{Person Split}
Table \ref{tab:results_person} presents the results for the parody prediction models with the data split by person. We observe the architectures using pre-trained text encoders (i.e. ULMFit, BERT, RoBERTa and XLNet) outperform both neural (BiLSTM-Att) and feature-based (LR-BOW and LR-BOW+POS) by a large margin across metrics with transformer architectures (BERT, RoBERTa and XLNet) performing best. The highest scoring model, RoBERTa, classifies accounts (parody and real) with an accuracy of 90, which is more than 8\% greater than the best non-transformer model (the ULMFit method). RoBERTa also outperforms the Logistic Regression baselines (LR-BOW and LR-BOW+POS) by more than 16 in accuracy and 13 in F1 score. Furthermore, it is the only model to score higher than 90 on precision.

\subsection{Gender Split}
Table \ref{tab:results_gender} shows the F1-scores obtained when training on the gender splits, i.e. training on male and testing on female accounts and vice versa. We first observe that models trained on the male set are in general more accurate than models trained on the female set, with the sole exception of ULMFit. This is probably due to the fact that the data set is imbalanced towards men as shown in Table \ref{tab:gender_split} (see also Section~\ref{sec:data}). We also do not observe a dramatic performance drop compared to the mixed-gender model on the person split (see Table~\ref{tab:results_person}). Again, RoBERTa is the most accurate model when trained in both splits, obtaining an F1-score of 87.11 and 84.87 for the male and female data respectively. The transformer-based architectures are again the best performing models overall, but the difference between them and the feature-based methods is smaller than it was on the person split.

\renewcommand{\arraystretch}{1.1}
\begin{table}[t!]
\centering
\begin{tabular}{|
>{\columncolor[HTML]{EFEFEF}}l |c|c|}
\hline
\multicolumn{3}{|c|}{\cellcolor[HTML]{C0C0C0}\textbf{Gender}}\\
\hline
\cellcolor[HTML]{C0C0C0}\textbf{Model} & \multicolumn{1}{l|}{\cellcolor[HTML]{C0C0C0}\textbf{M$\to$F}} & \multicolumn{1}{l|}{\cellcolor[HTML]{C0C0C0}\textbf{F$\to$M}}  \\\hline
 LR-BOW     & 78.89 &  76.63\\
 LR-BOW+POS & 78.74 &  76.74\\ \hline
 BiLSTM-Att & 77.00 &  77.11\\
 ULMFit     & 81.20 &  82.53\\
 BERT       & 85.85 & 84.40\\
 RoBERTa    & \textbf{87.11} & \textbf{84.87}\\
 XLNet      & 85.69 & 84.16\\
\hline
\end{tabular}
\caption{F1-scores for parody prediction splitting by gender (Male-M, Female-F). Best results are in bold.}
\label{tab:results_gender}
\end{table}

\subsection{Location Split}
Table \ref{tab:results_location} shows the F1-scores obtained training our models on the location splits: (i) train/dev on UK and RoW, test on US; (ii) train/dev on US and RoW, test on UK; and (iii) train/dev on US and UK, test on RoW. In general, the best results are obtained by training on the US \& UK split, while results of the models trained on the RoW \& US, and RoW \& UK splits are similar. The model with the best performance trained on US \& UK, and RoW \& UK splits is RoBERTa with F1 scores of 87.70 and 85.99 respectively. XLNet performs slightly better than RoBERTa when trained on RoW \& US data split. 

\renewcommand{\arraystretch}{1.3}
\begin{table}[t!]
\centering
\resizebox{\columnwidth}{!}{%
\begin{tabular}{|
>{\columncolor[HTML]{EFEFEF}}l |c|c|c|}
\hline
\multicolumn{4}{|c|}{\cellcolor[HTML]{C0C0C0}\textbf{Location}}\\
\hline
\cellcolor[HTML]{C0C0C0}\textbf{Model} & \multicolumn{1}{l|}{\cellcolor[HTML]{C0C0C0}\textbf{\includegraphics[scale=0.9]{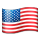} + \includegraphics[scale=0.9]{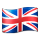} $\to$ \includegraphics[scale=0.9]{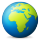}}} &
\multicolumn{1}{l|}{\cellcolor[HTML]{C0C0C0}\textbf{\includegraphics[scale=0.9]{emoji_images/hires/earth.pdf} + \includegraphics[scale=0.9]{emoji_images/hires/us.pdf} $\to$ \includegraphics[scale=0.9]{emoji_images/hires/uk.pdf}}} &
\multicolumn{1}{l|}{\cellcolor[HTML]{C0C0C0}\textbf{\includegraphics[scale=0.9]{emoji_images/hires/earth.pdf} + \includegraphics[scale=0.9]{emoji_images/hires/uk.pdf} $\to$ \includegraphics[scale=0.9]{emoji_images/hires/us.pdf}}}  \\\hline
 LR-BOW     & 78.58 & 78.27 & 77.97\\
 LR-BOW+POS & 78.27 & 77.88 & 78.08\\ \hline
 BiLSTM-Att & 80.29 & 77.59 & 73.19 \\
 ULMFit     & 83.47 & 81.55 & 81.55\\
 BERT       & 86.69 & 83.78 & 83.12\\
 RoBERTa    & \textbf{87.70} & \textbf{85.10} & \textbf{85.99}\\
 XLNet      & 85.32 & \textbf{85.17} & 85.32\\
\hline
\end{tabular}
}
\caption{F1-scores for parody prediction splitting by location. Best results are in bold.}
\label{tab:results_location}
\end{table}

\subsection{Discussion}
Through experiments over three different data splits, we show that all models predict parody tweets consistently above random, even if tested on people unseen in training. In general, we observe that the pre-trained contextual embedding based models perform best, with an average of around 10 F1 better than the linear methods. From these methods, we find that RoBERTa outperforms the other methods by a small, but consistent margin, similar to past research~\cite{roberta}. Further, we see that the predictions are robust to any location or gender specific differences, as the performance on held-out locations and genders are close to when splitting by person with a maximum of $<5$ F1 drop, also impacted by training on less data (e.g. female users). This highlights the fact that our models capture information beyond topics or features specific to any person, gender or location and can potentially identify stylistic differences between parody and real tweets. 

\section{Analysis}
\label{sec:analysis}

We finally perform an analysis based on our novel data set to uncover the peculiarities of political parody and  understand the limits of the predictive models.

\subsection{Linguistic Feature Analysis}


\begin{table}[!t]
\small
\centering
\begin{tabular}{|l|l|l|l|}
\hline
\rowcolor[HTML]{9B9B9B} 
\multicolumn{2}{|c|}{\cellcolor[HTML]{9B9B9B}Real}                                                     & \multicolumn{2}{c|}{\cellcolor[HTML]{9B9B9B}Parody}                                                   \\ \hline
\rowcolor[HTML]{C0C0C0} 
\multicolumn{1}{|c|}{\cellcolor[HTML]{C0C0C0}Feature} & \multicolumn{1}{c|}{\cellcolor[HTML]{C0C0C0}r} & \multicolumn{1}{c|}{\cellcolor[HTML]{C0C0C0}Feature} & \multicolumn{1}{c|}{\cellcolor[HTML]{C0C0C0}r} \\ \hline
\rowcolor[HTML]{C0C0C0} 
\multicolumn{4}{|l|}{\cellcolor[HTML]{C0C0C0}Unigrams}                                                                                                                                                         \\ \hline
our                                                   & 0.140                                          & i                                                    & 0.181                                          \\ \hline
in                                                    & 0.131                                          & ?                                                    & 0.156                                          \\ \hline
and                                                   & 0.129                                          & \textless{}mention\textgreater{}                     & 0.145                                          \\ \hline
:                                                     & 0.118                                          & me                                                   & 0.136                                          \\ \hline
\&                                                    & 0.114                                          & not                                                  & 0.106                                          \\ \hline
today                                                 & 0.105                                          & like                                                 & 0.097                                          \\ \hline
to                                                    & 0.105                                          & my                                                   & 0.095                                          \\ \hline
of                                                    & 0.098                                          & dude                                                 & 0.094                                          \\ \hline
the                                                   & 0.091                                          & don't                                                & 0.090                                          \\ \hline
at                                                    & 0.087                                          & i'm                                                  & 0.087                                          \\ \hline
lhl                                                   & 0.086                                          & just                                                 & 0.083                                          \\ \hline
great                                                 & 0.085                                          & know                                                 & 0.081                                          \\ \hline
with                                                  & 0.084                                          & \#feeltheburp                                        & 0.078                                          \\ \hline
de                                                    & 0.079                                          & you                                                  & 0.076                                          \\ \hline
meeting                                               & 0.078                                          & \#callmedick                                         & 0.075                                          \\ \hline
for                                                   & 0.077                                          & \#imwithme                                           & 0.073                                          \\ \hline
across                                                & 0.073                                          & "                                                    & 0.073                                          \\ \hline
families                                              & 0.073                                          & \#visionzero                                         & 0.069                                          \\ \hline
on                                                    & 0.070                                          & if                                                   & 0.069                                          \\ \hline
country                                               & 0.067                                          & have                                                 & 0.067                                          \\ \hline
\rowcolor[HTML]{C0C0C0} 
\multicolumn{4}{|l|}{\cellcolor[HTML]{C0C0C0}POS (Unigrams and Bigrams)}                                                                                                                                        \\ \hline
NN IN                                                 & 0.1600                                         & RB                                                   & 0.1749                                         \\ \hline
IN                                                    & 0.1507                                         & PRP                                                  & 0.1546                                         \\ \hline
CC                                                    & 0.1309                                         & RB VB                                                & 0.1271                                         \\ \hline
IN JJ                                                 & 0.1210                                         & VBP                                                  & 0.1206                                         \\ \hline
NNS IN                                                & 0.1165                                         & VBP RB                                               & 0.1123                                         \\ \hline
NN CC                                                 & 0.1114                                         & .                                                    & 0.1114                                         \\ \hline
IN NN                                                 & 0.1048                                         & NNP NNP                                              & 0.1094                                         \\ \hline
NN TO                                                 & 0.1030                                         & NN NNP                                               & 0.1057                                         \\ \hline
NNS TO                                                & 0.1013                                         & WRB                                                  & 0.0925                                         \\ \hline
TO                                                   & 0.1001                                         & VBP PRP                                              & 0.0904                                         \\ \hline
CC JJ                                                 & 0.0972                                         & IN PRP                                               & 0.0890                                         \\ \hline
IN DT                                                 & 0.0941                                         & NN VBP                                               & 0.0863                                         \\ \hline
: JJ                                                  & 0.0875                                         & RB .                                                 & 0.0854                                         \\ \hline
NNS                                                   & 0.0855                                         & NNP                                                  & 0.0837                                         \\ \hline
: NN                                                  & 0.0827                                         & JJ VBP                                               & 0.0813                                         \\ \hline
\end{tabular}
\caption{Feature correlations with parody and real tweets, sorted by Pearson correlation (r). All correlations are significant at $p < .01$, two-tailed t-test. }
\label{tab:linguisticAnalysis}
\end{table}

We first analyse the linguistic features specific of real and parody tweets. For this purpose, we use the method introduced in~\cite{schwartz2013personality} and used in several other analyses of user traits~\cite{preoctiuc2017beyond} or speech acts~\cite{preotiuc-pietro-etal-2019-automatically}. We thus rank the feature sets described in Section~\ref{sec:models} using univariate Pearson correlation (note that for the analysis we use POS tags instead of POS n-grams). Features are normalized to sum up to unit for each tweet. Then, for each feature, we compute correlations independently between its distribution across posts and the label of the post (parody or not). 

Table \ref{tab:linguisticAnalysis} presents the top unigrams and part-of-speech features correlated with real and parody tweets.
We first note that the top features related to either parody or genuine tweets are function words or related to style, as opposed to the topic. This enforces that the make-up of the data set or any of its categories are not impacted by topic choice and parody detection is mostly a stylistic difference. The only exception are a few hashtags related to parody accounts (e.g. \#imwithme), but on a closer inspection, all of these are related to tweets from a single parody account and are thus not useful in prediction by any setup, as tweets containing these will only appear in either the train or test set.

The top features related to either category of tweets are pronouns (`our' for genuine tweets, `i' for parody tweets). In general, we observe that parody tweets are much more personal and include possessives (`me', `my', `i', ``i'm'', PRP) or second person pronouns (`you'). This indicates that parodies are more personal and direct, which is also supported by use of more @-mentions and quotation marks. The real politician tweets are more impersonal and the use of `our' indicates a desire to include the reader in the conversation.

The real politician tweets include more stopwords (e.g. prepositions, conjunctions, determiners), which indicate that these tweets are more well formed. Conversely, the parody tweets include more contractions (``don't'', ``i'm''), hinting to a less formal style (`dude'). Politician tweets frequently use their account to promote events they participate in or are relevant to the day-to-day schedule of a politician, as hinted by several prepositions (`at', `on') and words (`meeting', ``today')~\cite{preotiuc-pietro-devlin-marier-2019-analyzing}.
For example, this is a tweet of the U.S. Senator from Connecticut, Chris Murphy: %
\begin{quote}
\small
\it
    Rudy Giuliani is in Ukraine \textbf{today}, \textbf{meeting} with Ukranian leaders on behalf of the President of the United States, representing the President's re-election campaign.[...]




\end{quote}
%

Through part-of-speech patterns, we observe that parody accounts are more likely to use verbs in the present singular (VBZ, VBP). This hints that parody tweets explicitly try to mimic direct quotes from the parodied politician in first person and using present tense verbs, while actual politician tweets are more impersonal. Adverbs (RB) are used predominantly in parodies and a common sequence in parody tweets is adverbs followed by verbs (RB VB) which can be used to emphasize actions or relevant events. For example, the following is a tweet of a parody account ($@$Queen\_Europe) of Angela Merkel:
\begin{quote}
\small
\it
    I mean, the Brexit Express \textbf{literally appears} to be going backwards but OK $<$url$>$
\end{quote}
\subsection{Error Analysis}

Finally, we perform an error analysis to examine the behavior of our best performing model (RoBERTa) and identify potential limitations of the current approaches. The first example is a tweet by the former US president Barack Obama which was classified as parody while it is in fact a real tweet:
\begin{quote}
\small
\it
Summer's almost over, Senate Leaders. \#doyourjob $<$url$>$
\end{quote}

\noindent Similarly, the next tweet was posted by the real account of the Virginia governor, Ralph Northam:
\begin{quote}
\small
\it
At this point, the list of Virginians Ed Gillespie *hasn't* sold out is shorter than the folks he has. $<$url$>$
\end{quote}

\noindent Both of the tweets above contain humoristic elements and come off as confrontational, aimed at someone else which is more prevalent in parody. We hypothesize that the model picked up this information to classify these tweets as parody. From the previous analyses, we noticed that tweets by real politicians often convey information in a more neutral or impersonal way. On the other hand, the following tweet was posted by a Mitt Romney parody account and was classified as real:


\begin{quote}
\small
\it
    It's up to you, America: do you want a repeat of the last four years, or four years staggeringly worse than the last four years?
\end{quote}

This parody tweet, even though it is more opinionated, is more similar in style to a slogan or campaign speech and is therefore missclassified. 
Lastly, the following is a tweet from former President Obama that was misclassified as parody:
\begin{quote}
\small
\it
    It's the $\#$GimmeFive challenge, presidential style. $<$url$>$
\end{quote}
The reason behind is that there are politicians, such as Barack Obama, who often write in an informal manner and this may cause the models to misclassify this kind of tweets. 

\section{Conclusion}

We presented the first study of parody using methods from computational linguistics and machine learning, a related but distinct linguistic phenomenon to irony and sarcasm. Focusing on political parody in social media, we introduced a freely available large-scale data set containing a total of 131,666 English tweets from 184 real and corresponding parody accounts. We defined parody prediction as a new binary classification task at a tweet level and evaluated a battery of feature-based and neural models achieving high predictive accuracy of up to 89.7\% F1 on tweets from people unseen in training.
%
%

In the future, we plan to study more in depth the stylistic and figurative devices used for parody, extend the data set beyond the political case study and explore human behavior regarding parody, including how this is detected and diffused through social media. 

\section*{Acknowledgments}
We thank Bekah Hampson for providing early input and helping with the data annotation. NA is supported by ESRC grant ES/T012714/1 and an Amazon AWS Cloud Credits for Research Award.

\clearpage

 \bibliography{library}
 \bibliographystyle{acl_natbib}

\end{document}